# Elicitation of Probabilities for Belief Networks: Combining Qualitative and Quantitative Information

**Marek J. Druzdzel**
University of Pittsburgh
Department of Information Science
and Intelligent Systems Program
Pittsburgh, PA 15260, U.S.A.
*marek@lis.pitt.edu*

**Linda C. van der Gaag**
Utrecht University
Department of Computer Science
P.O. Box 80.089
3508 TB Utrecht, The Netherlands
*linda@cs.ruu.nl*

## Abstract

Although the usefulness of belief networks for reasoning under uncertainty is widely accepted, obtaining numerical probabilities that they require is still perceived a major obstacle. Often not enough statistical data is available to allow for reliable probability estimation. Available information may not be directly amenable for encoding in the network. Finally, domain experts may be reluctant to provide numerical probabilities. In this paper, we propose a method for elicitation of probabilities from a domain expert that is non-invasive and accommodates whatever probabilistic information the expert is willing to state. We express all available information, whether qualitative or quantitative in nature, in a canonical form consisting of (in)equalities expressing constraints on the hyperspace of possible joint probability distributions. We then use this canonical form to derive second-order probability distributions over the desired probabilities.

## 1 INTRODUCTION

As the increasing number of successful applications demonstrate, *belief networks* [Pearl, 1988] have by now established their position of valuable representations of uncertainty in Artificial Intelligence (AI) research. A belief network (also referred to as *probabilistic network* or *causal network*) consists of a qualitative part, encoding a domain's variables and the probabilistic influences among them in a directed graph, and a quantitative part, encoding probabilities over these variables. Building the qualitative part of a belief network has parallels to other AI approaches and, although it may require significant effort, generally is not considered the hardest part in belief network construction. In most cases this task is dominated by the task of acquiring the quantification of the network.

Quantifying a belief network amounts to assessing probability distributions for each of the network's variables conditional on their direct predecessors in the directed graph. In most domains, at least some information is available to this end, be it from literature or from domain experts. However, this information often is not directly amenable to encoding in a belief network. For example, available information may not be numerical in nature. An expert may be certain of the fact that some values of a statistical variable $A$ make some values of a variable $B$ more likely, and perhaps have an idea of the lower and upper bounds on the numerical strength of this influence, yet may not be able to give exact numbers. Also, available probabilities may not match the probabilities to be assessed. Medical literature, for example, often reports probabilities of symptoms given diseases but usually not the probabilities of symptoms given no diseases and not necessarily the specific probabilities required for the intermediate disease states modeled in the network. Moreover, experts may feel more confident providing estimates of conditional probabilities in the diagnostic direction than in the causal direction of probabilistic influence.

Probabilistic information is available in many different shapes. It ranges from numerical point and interval probabilities, through order of magnitude estimates and signs of influences and synergies, to purely qualitative statements concerning independence of variables. This range has inspired a variety of schemes for reasoning under uncertainty. Some of these schemes build on quantitative information such as belief networks [Pearl, 1988] and undirected graphical models [Whittaker, 1990]; others build on partial numerical specifications, allowing for interval rather than point probabilities [Breese and Fertig, 1991; Coletti *et al.*, 1991; Coletti, 1994; van der Gaag, 1991] or for order of magnitude estimates [Goldszmidt and Pearl, 1992]. Yet other schemes are purely qualitative in nature, such as qualitative probabilistic networks [Wellman, 1990]. Also non-probabilistic schemes have been proposed, each addressing a specific type of uncertainty, such as Dempster-Shafer theory [Shafer, 1976], possibility theory [Zadeh, 1978], and non-monotonic logics [Pearl, 1989]. Each of these schemes typically allows for en-coding only a few types of information. A unifying principle that would allow combining the various types of information has been lacking so far, making it hard to utilize the variety of information available in practice.

With the purpose of quantifying belief networks in mind, we propose a method for accommodating both qualitative and quantitative probabilistic information about a yet *unknown* joint probability distribution Pr over a set of variables $V$. The basic idea of our method is to consider the *distribution hyperspace* of all possible joint probability distributions over $V$. The true, yet unknown distribution Pr is a point in this hyperspace. If no information is available about Pr, then the true distribution can be any point in the distribution hyperspace. Information about Pr, whether qualitative or quantitative, expresses a constraint on the hyperspace since certain distributions become incompatible with this information. Probability elicitation can now be looked upon as constraining the distribution hyperspace as much as possible. To this end, we express all probabilistic information that is available about the unknown distribution as constraints. Assuming that all joint probability distributions that are compatible with the available information are equally likely, we then derive second-order probability distributions over the probabilities to be assessed. These second-order distributions may be used directly or may be a starting point for further refinement. Note that our approach provides a common denominator for various types of probabilistic information. Also note that by interpreting the qualitative and quantitative information that a domain expert is willing to state, we effectively provide for *non-invasive* elicitation of probabilities. We believe that our method is a valuable supplement to the classical decision-analytic techniques of probability elicitation.

The remainder of this paper is structured as follows. Section 2 introduces a simple belief network that will be used throughout the paper and gives examples of probabilistic information that is typically available for quantifying a network. Section 3 presents a canonical form for representing probabilistic information and Section 4 describes interpretation of various types of information within this canonical form. Section 5 demonstrates how information expressed in canonical form can be used to derive second-order probability distributions over probabilities of interest. We finish with a discussion and an outline of directions for further research in Section 6.

## 2  AN EXAMPLE

Consider building a highly simplified belief network modeling causes of HIV virus infection. Our network includes four variables: *HIV infection* ($H$), *needle sharing* ($N$), *sexual intercourse* ($I$), and *use of a condom* ($C$). We assume, for the sake of simplicity, that these variables are binary; for example $H$ has two outcomes, denoted $h$ and $\overline{h}$, representing "HIV infection present" and "HIV infection absent," respectively.

The first step in building a belief network is to design its structure in terms of probabilistic influences among its variables. Belief networks achieve clarity and large savings in terms of storage of a joint probability distribution by explicit representation of the independences holding among its variables. These independences are encoded in a directed acyclic graph, where each node represents a variable and each arc represents, informally speaking, a direct probabilistic influence between its incident nodes. Absence of an arc between two variables means that these variables do not influence each other directly, and hence are (conditionally) independent. For orienting the arcs in the graph, it is generally considered good practice to reflect the causal mechanisms [Druzdzel and Simon, 1993] of the domain. In our example, we may reasonably assume that sharing needles and condom usage are independent. Similarly, whether or not a person shares needles may be assumed independent of whether this person engages in sexual intercourse. One possible graph reflecting our beliefs concerning HIV infection is shown in Figure 1.

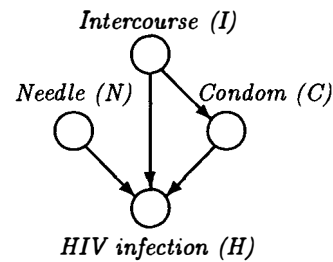

Figure 1: An example belief network for HIV infection.

Once the qualitative part of a network is considered robust, the network is quantified. To this end, for each variable the probabilities of its values conditional on the values of its direct predecessors in the graph have to be assessed. For the graph shown in Figure 1, numbers representing $\Pr(N)$, $\Pr(I)$, $\Pr(C|I)$, and $\Pr(H|NIC)$ are required. Obtaining these numbers is considered to be far more difficult than configuring the qualitative part of the network, mainly because of difficulties in obtaining statistical data and in eliciting probabilities from domain experts. In our example, there are several sources of information that can help in obtaining the required probabilities. Morbidity tables may provide $\Pr(h)$, a point estimate of the prevalence of HIV in the population of interest. We may get ball-park estimates on frequencies of sexual intercourse and condom usage in intercourse, that is, $\Pr(i)$ and $\Pr(c|i)$. We further know that condoms are used primarily during intercourse, so $\Pr(\overline{i}|c)$ is close to zero. In addition, various populations of intravenous drug users have been studied with respect to their needle sharing habits. Findings from these studies may help in assessing $\Pr(n)$. Also, statistics may be obtained concerning the way of contracting

HIV from among the infected population, yielding estimates for $\Pr(n|h)$ and $\Pr(i|h)$, or perhaps even for $\Pr(ic|h)$ and $\Pr(i\bar{c}|h)$. There is also semi-numerical information available. For example, the probability of contracting HIV by needle sharing is higher than the probability of contracting it in sexual intercourse, that is, $\Pr(h|n) > \Pr(h|i)$. Also, the relatively small number of intravenous drug users compared to the size of the sexually active population suggests that $\Pr(i) > \Pr(n)$.

Besides (semi-)numerical information, we have a body of qualitative information on the subject. We are quite certain that both sharing a needle and a sexual intercourse with an HIV carrier make infection more likely. We know that using a condom during an intercourse decreases the likelihood of contracting HIV. These two pieces of information express *qualitative influences* between pairs of variables. A formal interpretation of qualitative influences has been proposed by [Wellman, 1990] in terms of statistical dominance. This property is also useful in capturing qualitative synergies between variables. A positive (negative) *additive synergy* [Wellman, 1990] captures the property that the joint influence of two variables on a third variable is larger (smaller) than the sum of their individual influences. In our example, condom usage and sexual intercourse are negatively additively synergistic: using a condom diminishes the influence of having intercourse on contracting HIV. *Product synergy* [Druzdzel and Henrion, 1993; Henrion and Druzdzel, 1991; Wellman and Henrion, 1993], on the other hand, captures intercausal interaction. An example is the negative intercausal interaction known as "explaining away" [Pearl, 1988] which models negative influence of the presence of one cause on the likelihood of another cause being present given an observed common effect. In our example, needle sharing and sexual intercourse are negatively product synergistic: given HIV infection, factual knowledge about needle sharing reduces the likelihood of intercourse being the cause of the infection.

These examples demonstrate that practical domains offer a wealth of probabilistic information which, although not always in the shape of numbers that are directly amenable to encoding in a belief network, may facilitate assessing the required probabilities.

## 3 CANONICAL FORM

Our canonical form for interpreting probabilistic information builds on the property that any joint probability distribution on a set of variables $V$ is uniquely defined by the probabilities of all possible combinations of values for all variables from $V$. If these probabilities are known, then any (other) probability from the distribution can be computed from them by applying the basic rules of marginalization and conditioning from probability theory. We will call combinations of values for all variables *constituent assignments*. The probabilities of constituent assignments in a joint probability distribution will be called its *constituent probabilities*. The set of all possible joint probability distributions on $V$ now can be looked upon as spanning a hyperspace whose dimensions correspond with constituent probabilities.

Any information about the true, yet unknown probability distribution Pr can now be represented as a system of (in)equalities involving this distribution's constituent probabilities as unknowns. Any solution to this system of (in)equalities is a joint probability distribution that is compatible with the available information. If the system has a unique solution, then the information provided suffices for uniquely defining Pr [van der Gaag, 1991]. Note that in case the system does not have any solution at all, the information about the unknown distribution Pr is inconsistent. This view of probability is largely based on the early work by Boole [Boole, 1958] on the foundations of probability theory.

We introduce some notational conventions. We take $V = \{V_1, \ldots, V_n\}$, $n \geq 1$, to be a set of variables, where each variable $V_i$ can take one of $k_i$ values. We will use $v_{i_j}$ to denote $V_i$ taking the $j$-th value from its domain, $j = 1, \ldots, k_i$. Note that the set of all constituent assignments for $V$ comprises $k = \prod_{i=1,\ldots,n} k_i$ elements.

Now, consider an assignment $b$ for an arbitrary subset of variables from $V$ and its unknown probability $\Pr(b)$. The assignment $b$ can be written as a disjunction of constituent assignments $c_i$ using basic logical laws. In fact, here exists a unique set of indices $I_b \subseteq \{1, \ldots, k\}$, called the *index set* for $b$, such that $b = \bigvee_{i \in I_b} c_i$. Since all constituent assignments are mutually exclusive, the probability $\Pr(b)$ can be expressed as the sum of the probabilities of the constituent assignments $b$ is built from. So, from $\Pr(b) = \sum_{i \in I_b} \Pr(c_i)$ we find that $\Pr(b)$ can be expressed as

$$d_1 x_1 + d_2 x_2 + \cdots + d_k x_k \qquad (1)$$

where $x_i = \Pr(c_i)$, $i = 1, \ldots, k$, and $d_i = 1$ if $i \in I_b$ and $d_i = 0$ otherwise.

**Example:** Consider the example belief network for HIV infections from Section 2. There are sixteen constituent assignments for the variables involved; an ordered list of these assignments is shown in Table 1. Now consider the assignment expressing a person's having sexual intercourse without using a condom, that is, the assignment $i\bar{c}$. This assignment can be written as

$$\begin{aligned} i\bar{c} &= hni\bar{c} \lor \bar{h}ni\bar{c} \lor h\bar{n}i\bar{c} \lor \bar{h}\bar{n}i\bar{c} \\ &= c_5 \lor c_8 \lor c_{10} \lor c_{13} \end{aligned}$$

Note that the index set $I_{i\bar{c}}$ equals $I_{i\bar{c}} = \{5, 8, 10, 13\}$. The probability $\Pr(i\bar{c})$ can now be expressed as

$$\begin{aligned} \Pr(i\bar{c}) &= \Pr(c_5) + \Pr(c_8) + \Pr(c_{10}) + \Pr(c_{13}) \\ &= x_5 + x_8 + x_{10} + x_{13} \end{aligned}$$

$$\begin{array}{llll}
c_1 = hnic & c_5 = hni\bar{c} & c_9 = h\bar{n}ic & c_{13} = \bar{h}\bar{n}i\bar{c} \\
c_2 = \bar{h}nic & c_6 = \bar{h}\bar{n}ic & c_{10} = h\bar{n}i\bar{c} & c_{14} = \bar{h}n\bar{i}\bar{c} \\
c_3 = h\bar{n}ic & c_7 = \bar{h}n\bar{i}c & c_{11} = hn\bar{i}\bar{c} & c_{15} = h\bar{n}\bar{i}\bar{c} \\
c_4 = hn\bar{i}c & c_8 = \bar{h}ni\bar{c} & c_{12} = \bar{h}\bar{n}i\bar{c} & c_{16} = \bar{h}\bar{n}\bar{i}\bar{c}
\end{array}$$

Table 1: Constituent assignments for the HIV belief network.

Note that in terms of expression (1), we have that $d_5 = d_8 = d_{10} = d_{13} = 1$ and $d_i = 0$ for all $i \neq 5, 8, 10, 13$.
□

Posterior probabilities are expressed in canonical form in a similar way. Consider a posterior probability $\Pr(b_1|b_2)$ where $b_1, b_2$ denote assignments for sets of variables. From $\Pr(b_1|b_2) = \frac{\Pr(b_1 b_2)}{\Pr(b_2)}$, we have that $\Pr(b_1|b_2)$ can be expressed as

$$\frac{d_{1,1}x_1 + d_{2,1}x_2 + \cdots + d_{k,1}x_k}{d_{1,2}x_1 + d_{2,2}x_2 + \cdots + d_{k,2}x_k}$$

where $x_i = \Pr(c_i)$, and $d_{i,1} = 1$ if $i \in I_{b_1 b_2}$ and $d_{i,1} = 0$ otherwise, and $d_{i,2} = 1$ if $i \in I_{b_2}$ and $d_{i,2} = 0$ otherwise. Note that $d_{i,2} = 1$ whenever $d_{i,1} = 1$.

## 4 INTERPRETATION OF PROBABILISTIC INFORMATION

In this section, we address expressing axiomatic information, point estimates, probability intervals, comparisons, qualitative influences, and additive synergies in our canonical form. We have designed similar expressions for other types of information, such as independences, order of magnitude estimates, product synergies, and noisy-OR gates. A technical report providing all interpretations is in preparation.

### 4.1 AXIOMATIC INFORMATION

Even if no specific information is available about an unknown joint probability distribution, there still is probabilistic information that holds for any distribution. This information concerns the basic axiomatic properties of a joint probability distribution.

The unknown joint probability distribution Pr is known to be *normed*, that is, $\Pr(true) = 1$. This property is expressed in canonical form by the equality

$$x_1 + \cdots + x_k = 1 \qquad (2)$$

where $x_i = \Pr(c_i)$, $i = 1, \ldots, k$.

Also, the probability $\Pr(b)$ for any assignment $b$ of a set of variables from $V$ is known to be a *non-negative* real number. More in specific, we have that for any constituent probability $\Pr(c_i)$, $i = 1, \ldots, k$, the property $\Pr(c_i) \geq 0$ holds. This information is expressed in canonical form in $k$ inequalities of the form

$$x_i \geq 0 \qquad (3)$$

for $i = 1, \ldots, k$. Note that if all constituent probabilities are non-negative, then all other probabilities are non-negative as well. Hence, there is no need to specify any additional constraints for this information. Also, note that the constraints (2) and (3) imply that $\Pr(b) \leq 1$ for any assignment $b$.

### 4.2 POINT PROBABILITIES, INTERVALS, AND COMPARISONS

A *point estimate* for a prior probability is a statement of the form $\Pr(b) = p$, $0 \leq p \leq 1$, where $b$ is an assignment for an arbitrary subset of variables. Let $I_b$ be the index set for $b$. Then, the point estimate is expressed in canonical form as

$$d_1 x_1 + \cdots + d_k x_k = p$$

where $x_i = \Pr(c_i)$, $i = 1, \ldots, k$, and $d_i = 1$ if $i \in I_b$ and $d_i = 0$ otherwise.

**Example:** Consider once more the HIV belief network. The prevalence of HIV infection in the U.S. population is $\Pr(h) = 0.005$ according to morbidity tables. This information is expressed in canonical form as

$$x_1 + x_3 + x_4 + x_5 + x_9 + x_{10} + x_{11} + x_{15} = 0.005$$

□

A point estimate for a posterior probability is a statement of the form $\Pr(b_1|b_2) = p$, $0 \leq p \leq 1$, where $b_1, b_2$ denote assignments for sets of variables. From $\Pr(b_1|b_2) = \frac{\Pr(b_1 b_2)}{\Pr(b_2)}$, we have that $\Pr(b_1 b_2) = p \cdot \Pr(b_2)$, and therefore $\Pr(b_1 b_2) - p \cdot \Pr(b_2) = 0$. The probabilities $\Pr(b_1 b_2)$ and $\Pr(b_2)$ now are expressed in terms of constituent probabilities as before. The point estimate for $\Pr(b_1|b_2)$ further indicates that $\Pr(b_2) > 0$ and, therefore, gives rise to yet another inequality in terms of constituent probabilities.

Similar expressions in canonical form are found for probability intervals and comparisons of probabilities. A *probability interval* is a statement expressing an upper and a lower bound on a prior or posterior probability. Such a statement may be of the form $p_1 \leq \Pr(b) \leq p_2$ where $b$ is an assignment for an arbitrary subset of variables and $p_1, p_2$ are real numbers such that $0 \leq p_1 < p_2 \leq 1$. A *comparison* between two prior probabilities can be of the form $a_1 \cdot \Pr(b_1) \leq a_2 \cdot \Pr(b_2)$ where $b_1, b_2$ are assignments for subsets of variables from $V$ and $a_1, a_2$ are (non-negative) real numbers. These statements are expressed in canonical form by writing the probabilities $\Pr(b)$, $\Pr(b_1)$, and $\Pr(b_2)$ in terms of constituent probabilities.

### 4.3 QUALITATIVE INFLUENCES

A *qualitative influence* is a symmetric property describing the sign of probabilistic interaction between two variables $V_1$ and $V_0$, and builds on an ordering of these variables' values. A positive qualitative influence from $V_1$ to $V_0$ expresses that choosing a higher value for $V_1$ makes higher values of $V_0$ more likely, regardless of the values of other variables. More formally [Wellman, 1990], we say that the variable $V_1$ *positively influences* the variable $V_0$, denoted by $S^+(V_1, V_0)$, iff for all values $v_{0_m}$ of $V_0$, for all pairs of distinct values $v_{1_i} > v_{1_j}$ of $V_1$, and for all possible assignments $b$ for the set of $V_0$'s direct predecessors other than $V_1$, we have

$$\Pr(V_0 \geq v_{0_m} | v_{1_i}, b) \geq \Pr(V_0 \geq v_{0_m} | v_{1_j}, b)$$

*Negative qualitative influence* and *zero qualitative influence* are defined analogously.

The statement $S^+(V_1, V_0)$ is expressed in canonical form by expressing a set of inequalities in this form. There is one inequality for each combination of one value $v_{0_m}$ of $V_0$, one pair of values $v_{1_i}, v_{1_j}$ of $V_1$, and one assignment $b$ of $V_0$'s other predecessors than $V_1$; this inequality expresses that

$$\sum_{l=m}^{k_0} \Pr(v_{0_l} | v_{1_i}, b) \geq \sum_{l=m}^{k_0} \Pr(v_{0_l} | v_{1_j}, b)$$

Note that there are $\binom{k_1}{2} \cdot (k_0 - 1) \cdot K$ such inequalities, where $K$ is the number of possible assignments for the set of direct predecessors of $V_0$ other than $V_1$. As these inequalities involve posterior probabilities, each of them gives rise to two additional inequalities.

**Example:** For quantifying our HIV belief network, the available information indicates that needle sharing positively influences HIV infection, that is, $S^+(N, H)$. This statement translates into the four inequalities:

$$\Pr(h|nic) \geq \Pr(h|\overline{n}ic)$$
$$\Pr(h|n\overline{i}c) \geq \Pr(h|\overline{n}\overline{i}c)$$
$$\Pr(h|ni\overline{c}) \geq \Pr(h|\overline{n}i\overline{c})$$
$$\Pr(h|n\overline{i}\overline{c}) \geq \Pr(h|\overline{n}\overline{i}\overline{c})$$

and eight additional inequalities expressing that $\Pr(nic) > 0, \ldots, \Pr(\overline{n}\overline{i}\overline{c}) > 0$. Note that the statement $S^+(N, H)$ gives rise to the total of twelve inequalities. The first inequality mentioned above is expressed in canonical form as

$$x_2 x_3 - x_1 x_6 \geq 0$$

The other inequalities are expressed analogously. □

### 4.4 QUALITATIVE SYNERGIES

An *additive synergy* pertains to the joint influence of two variables $V_1$ and $V_2$ on a third variable $V_0$, and, similarly to qualitative influence, builds on an ordering of these variables' values. A positive additive synergy of $V_1$ and $V_2$ with respect to $V_0$ expresses that the joint influence of $V_1$ and $V_2$ is greater than the sum of their individual influences. More formally [Wellman, 1990], we say that the variables $V_1$ and $V_2$ exhibit *positive additive synergy* with respect to $V_0$, denoted by $Y^+(\{V_1, V_2\}, V_0)$, iff for all values $v_{0_m}$ of $V_0$, for all pairs of values $v_{1_i} > v_{1_j}$ of $V_1$ and $v_{2_{i'}} > v_{2_{j'}}$ of $V_2$, and for all possible assignments $b$ for the set of $V_0$'s direct predecessors not including $V_1$ and $V_2$, we have

$$\Pr(V_0 \geq v_{0_m} | v_{1_i}, v_{2_{i'}}, b) + \Pr(V_0 \geq v_{0_m} | v_{1_j}, v_{2_{j'}}, b)$$
$$\geq \Pr(V_0 \geq v_{0_m} | v_{1_i}, v_{2_{j'}}, b) + \Pr(V_0 \geq v_{0_m} | v_{1_j}, v_{2_{i'}}, b)$$

*Negative additive synergy* and *zero additive synergy* are defined analogously.

The statement $Y^+(\{V_1, V_2\}, V_0)$ is expressed in canonical form by a set of inequalities in the above form. There is one inequality for each combination of one value $v_{0_m}$ of $V_0$, one pair of values $v_{1_i}, v_{1_j}$ of $V_1$, one pair of values $v_{2_{i'}}, v_{2_{j'}}$ of $V_2$, and one assignment $b$ of $V_0$'s other direct predecessors than $V_1$ and $V_2$; there are $\binom{k_1}{2} \cdot \binom{k_2}{2} \cdot (k_0 - 1) \cdot K$ such inequalities, where $K$ is the number of possible assignments for the set of direct predecessors of $V_0$ other than $V_1$ and $V_2$. As these inequalities involve posterior probabilities, each of them gives rise to additional inequalities as outlined before.

**Example:** Consider once more our HIV belief network under construction. The available information indicates that there is a negative additive synergy between sexual intercourse and using a condom with respect to HIV infection, that is, that $Y^-(\{I, C\}, H)$. This statement translates into the two inequalities:

$$\Pr(h|nic) + \Pr(h|n\overline{i}\overline{c}) \leq \Pr(h|ni\overline{c}) + \Pr(h|n\overline{i}c)$$
$$\Pr(h|\overline{n}ic) + \Pr(h|\overline{n}\overline{i}\overline{c}) \leq \Pr(h|\overline{n}i\overline{c}) + \Pr(h|\overline{n}\overline{i}c)$$

and eight additional inequalities expressing that $\Pr(nic) > 0, \ldots, \Pr(\overline{n}\overline{i}\overline{c}) > 0$. Note that the statement $Y^-(\{I, C\}, N)$ gives rise to the total of ten inequalities. The first inequality above leads to

$$-x_1 x_4 x_5 x_{14} - x_2 x_4 x_5 x_{11} - 2x_2 x_4 x_5 x_{14}$$
$$+x_1 x_5 x_7 x_{11} - x_2 x_5 x_7 x_{14} + x_1 x_4 x_8 x_{11}$$
$$-x_2 x_4 x_8 x_{14} + 2x_1 x_7 x_8 x_{11} + x_1 x_7 x_8 x_{14}$$
$$+x_2 x_7 x_8 x_{11} \leq 0$$

The other inequalities are expressed in canonical form analogously. □

*Product synergy* pertains to the interaction between two variables $V_1$ and $V_2$ conditional on their common descendant $V_0$ and expresses the sign of what is known as *intercausal influence* between $V_1$ and $V_2$. The most common type of product synergy is the negative product synergy, capturing the notion of "explaining away." We say that the variables $V_1$ and $V_2$ exhibit *negative product synergy* with respect to a particular value $v_{0_m}$ of variable $V_0$, written $X^-(\{V_1, V_2\}, v_{0_m})$, if for all pairs of values $v_{2_i} > v_{2_j}$ of $V_2$ and for all possible assignments $b$ for the set of $V_0$'s direct predecessors not including $V_1$ and $V_2$, we have

$$\Pr(V_1 \geq v_{1_i} | v_{2_i} v_{0_m} b) \leq \Pr(V_1 \geq v_{1_i} | v_{2_j} v_{0_m} b)$$

*Positive product synergy* and *zero product synergy* are defined analogously. Note that, in contrast to additive synergy, product synergy is defined with respect to *separate* values of the common effect $V_0$. There are, therefore, as many product synergies as there are values of $V_0$. A statement $X^-(\{V_1, V_2\}, v_{0_m})$ is expressed in canonical form much in the same way as qualitative influences and additive synergies.

It is worth noting that the above definition is considerably less complex than the definition proposed in [Druzdzel and Henrion, 1993]. The latter definition expresses product synergy in terms of the probability of $V_0$ conditional on $V_1$ and $V_2$ to allow for derivation of the sign of product synergy from an existing conditional distribution encoded in a network. In terms of the canonical form proposed in this paper, we can afford defining product synergy in terms of probability of $V_1$ conditional on $V_2$ and $V_0$. This does not have any effect on the interpretation of statements regarding product synergy yet simplifies the matters greatly.

## 5  ELICITATION OF PROBABILITIES

Our method for elicitation of probabilities from a domain expert amounts to reasoning about the information that is available about the unknown joint probability distribution. We have illustrated how various types of information are expressed in the canonical form as a system of (in)equalities with constituent probabilities as unknowns. This section shows how these (in)equalities can be used to derive second-order probability distributions over any probability of interest in the sense suggested by [Pearl, 1988].

### 5.1 DERIVATION OF SECOND-ORDER DISTRIBUTIONS

From the system of (in)equalities resulting from expression of available probabilistic information in canonical form, we can compute upper and lower bounds on any probability of interest. The length of a computed interval then indicates the uncertainty in the probability's value and hence is a measure for the incompleteness of the available information. This method has been proposed before by Van der Gaag in view of systems of linear (in)equalities [van der Gaag, 1991]. For probability elicitation, this method has the disadvantage that upper and lower bounds on a probability give insufficient insight into how likely a value from the interval is to be the actual probability. Nor do these bounds provide an estimate of the expected value of the probability. We would like to note that for decision making in presence of uncertainty about a probability $p$, knowing the expected value of $p$ suffices, even if the distribution over $p$ is unknown [Howard, 1988].

To yield insight in the likelihood of values for the true probability, and in particular to be able to derive its expected value, we propose using sampling to find second-order distributions for the probabilities to be assessed. For computing these second-order distributions, we randomly select points from the distribution hyperspace, assuming that all points in the hyperspace are equally likely to be the true distribution. For each selected distribution, we verify its compatibility with all available information, that is, we verify if it is a solution to the system of (in)equalities derived from this information. All selected distributions matching the available information are collected and scored for the probabilities to be assessed; the result is a second-order distribution over each such probability. We would like to note that computing second-order distributions is computationally expensive as it involves generating and investigating joint probability distributions described by their constituent probabilities and the number of these constituent probabilities is exponential in the number of variables discerned.

**Example:** Consider once again the HIV infection example belief network. We have expressed the following probabilistic information about the four variables $H$, $N$, $I$, and $C$ in canonical form: $\Pr(i|c) = 1$, $\Pr(i) > \Pr(n)$, $\Pr(h|n) > \Pr(h|i)$, and the information that between 10% and 25% of HIV-infections are caused by needle sharing, that is, $0.1 \leq \Pr(n|h) \leq 0.25$. From this information, we derived second-order distributions for the various probabilities to be assessed for the network by selecting 10,000 matching joint probability distributions. The histograms of the samples obtained for $\Pr(i)$ and $\Pr(h|n\bar{i}\bar{c})$ are shown in Figure 2. When normalized, these histograms express a second order probability distribution over $\Pr(i)$ and $\Pr(h|n\bar{i}\bar{c})$. Note that the information from which we derived these distributions did not pertain directly to these probabilities. Another point that we would like to emphasize here is that knowledge of intervals would be useless as the probability $\Pr(h|n\bar{i}\bar{c})$, for example, spans over the entire interval between 0 and 1.   □

We have implemented our method for computing second-order distributions in Allegro Common Lisp on a Hewlett Packard workstation. Our implementation is just a prototype and has been created to serve illustrative purposes. As the implementation is straightforward, it is rather slow and therefore leaves much room for algorithmic improvement.

Especially when very restrictive information about the joint probability distribution is available, randomly selecting distributions from the hyperspace tends to yield a huge number of samples that are not compatible with the available information and therefore are not useful. To improve on the ratio of useful samples, we envision a *pre-processing* step prior to the selec-tion of distributions. In this step, a part of the hyperspace in which the true joint probability distribution definitely lies is identified. To this end, all linear (in)equalities from the system at hand are collected and a standard linear-programming technique is applied to compute upper and lower bounds on all constituent probabilities. The thus computed bounds are guaranteed to be sound: no point in the hyperspace outside these bounds can represent the unknown probability distribution. These bounds, however, may not be tight as there may be other, yet unconsidered information. Selecting distributions is now performed within the bounds yielded by the pre-processing step.

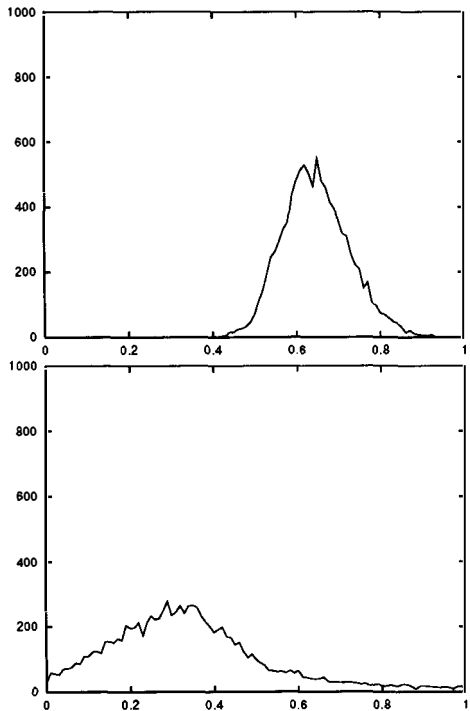

Figure 2: Histograms of the samples for Pr($i$) (upper) and Pr($h|n\bar{i}\bar{c}$) (lower).

### 5.2 FOCUSING ELICITATION

Reasoning about probabilistic information is computationally expensive. This is not surprising given that inference in belief networks is NP-hard [Cooper, 1990]. To allow for sidestepping the issue of complexity, we divide the problem of reasoning about qualitative and quantitative probabilistic information over all statistical variables in the network under construction into smaller subproblems and address these separately.

Division into subproblems is achieved by transforming the directed graph of the network into an undirected *chordal* graph that equally models independences from the distribution at hand. A chordal graph has the useful property that the joint probability distribution over the represented variables factorizes into marginal distributions on the separate cliques of this graph. This property allows for addressing the problem of elicitation of probabilities per clique. For transforming the directed graph of a belief network into a chordal graph, we make use of the transformation scheme designed by [Lauritzen and Spiegelhalter, 1988].

Computational complexity, however, is just one of the reasons for focusing elicitation of probabilities on small sets of variables. Focusing is also suggested by knowledge acquisition experience both in decision analysis and in expert systems design: human experts typically express information about short causal reasoning chains and feel uncomfortable when forced to provide more global information. An important property of the applied transformation is that, as for any variable and its direct predecessors a clique is yielded, causal mechanisms are never split up over different cliques and hence are never broken. We believe that the obtained cliques form small entities suitable for elicitation.

## 6 DISCUSSION

Although the usefulness of belief networks for representing and reasoning under uncertainty is widely accepted, eliciting probabilities for quantifying a network is often perceived a problem. It often turns out, however, that it is the need to express probabilistic information as *exact numbers* that tends to make domain experts feel uncomfortable: experts typically are able to state probabilistic information of a semi-numerical or qualitative nature with conviction and clarity, and hence with little cognitive effort. In this paper, we have proposed a method that allows for non-invasive elicitation of probabilities by interpreting and combining whatever an expert is willing to state. Our method can be used iteratively in the sense of starting the elicitation with only most robust and readily available information, and then narrowing down the focus of elicitation successively. As elicitation of probabilities from domain experts generally is a time-consuming and costly task, we expect this approach to lead to considerable savings. We believe that our method provides a valuable supplement to decision-analytic methods of probability elicitation.

Even though a non-invasive method of collecting information from experts may be less prone to conflicts than a method eliciting numerical probabilities, the constraints elicited may turn out to be inconsistent. Inconsistencies can arise from an expert's internal inconsistency or from disagreement among multiple experts and can occur either within a clique or between cliques. Detection of inconsitencies is quite straightforward. In accord with the decision analytic approach, we view inconsistencies as an additional opportunity to refine the elicitation by confronting the expert with conflicting statements. We believe that including both qualitative and quantitative statements in elicitation aids this refinement: qualitative information gener-ally is more robust and cognitively reliable. We plan to deal with inconsistencies by prioritizing the expert statements according to their expected robustness and suggesting the least robust constraints for revision. In the near future, we envision making our method the centerpiece of a general purpose computerized probability elicitation tool.

**Acknowledgements**

We thank one of the reviewers for references to the work of Coletti *et al.*